\documentclass[sigplan,10pt]{acmart} 
\usepackage{textcomp, libertine} 
\usepackage{import} 
\usepackage{graphicx} 
\graphicspath{{figs/}} 
\usepackage{tikz}
\usepackage[acronym,nowarn,hyperfirst=false]{glossaries}
\usepackage{algorithm} 
\usepackage{algpseudocode} 
\usepackage{subcaption}
\usepackage{listings}
\usepackage{dblfloatfix}    
\AtBeginDocument{%
  \providecommand\BibTeX{{%
    \normalfont B\kern-0.5em{\scshape i\kern-0.25em b}\kern-0.8em\TeX}}}
 
\copyrightyear{2021} 
\acmYear{2021}
\setcopyright{acmlicensed}\acmConference[EuroMLSys'21]{The 1st Workshop on Machine Learning and Systems}{April 26, 2021}{Online, United Kingdom}
\acmBooktitle{The 1st Workshop on Machine Learning and Systems (EuroMLSys'21), April 26, 2021, Online, United Kingdom}
\acmPrice{15.00}
\acmDOI{10.1145/3437984.3458841}
\acmISBN{978-1-4503-8298-4/21/04}
     
\usepackage{xcolor}

\definecolor{codegreen}{rgb}{0,0.6,0}
\definecolor{codegray}{rgb}{0.5,0.5,0.5}
\definecolor{codepurple}{rgb}{0.58,0,0.82}
\definecolor{backcolour}{rgb}{0.95,0.95,0.92}

\lstdefinestyle{mystyle}{
  backgroundcolor=\color{backcolour},   commentstyle=\color{codegreen},
  keywordstyle=\color{magenta},
  numberstyle=\tiny\color{codegray},
  stringstyle=\color{codepurple},
  basicstyle=\ttfamily\footnotesize,
  breakatwhitespace=false,         
  breaklines=true,                 
  captionpos=b,                    
  keepspaces=true,                 
  numbers=left,                    
  numbersep=5pt,                  
  showspaces=false,                
  showstringspaces=false,
  showtabs=false,                  
  tabsize=2
}

\newacronym{rl}{RL}{Reinforcement Learning}
\newacronym{bo}{BO}{Bayesian Optimization}
\newacronym{bnn}{BNN}{Bayesian Neural Network}
\newacronym{gp}{GP}{Gaussian Process}
\newacronym{bno}{BNO}{Bayesian Network Optimization}
\newacronym{acf}{ACF}{Acquisition function}
\newacronym{kv}{KV}{Key-Value}
\newacronym{dag}{DAG}{Directed Acyclic Graph}
\newacronym{ycsb}{YCSB}{Yahoo! Cloud Serving Benchmark}
\newacronym{ml}{ML}{Machine learning}
\newacronym{tpe}{TPE}{Tree-structured Parzen Estimator}
\newacronym{smac}{SMAC}{Sequential Model-Based Optimization}
\newacronym{hb}{HB}{Hyperband}
\newacronym{bohb}{BOHB}{Bayesian Optimization Hyperband}
\newacronym{pbt}{PBT}{Population Based Training}
\newacronym{ppo}{PPO}{Proximal Policy Optimization} 
\glsdisablehyper 
\begin{document}
\title{High-Dimensional Bayesian Optimization with Multi-Task Learning for RocksDB}
\author{Sami Alabed}
\email{sa894@cam.ac.uk}
\orcid{0000-0001-8716-526X}
\affiliation{%
  \institution{University of Cambridge}
  \city{Cambridge}
  \country{UK}
}
\author{Eiko Yoneki}
\email{eiko.yoneki@cl.cam.ac.uk}
\affiliation{%
  \institution{University of Cambridge}
  \city{Cambridge}
  \country{UK}
}
\renewcommand{\shortauthors}{Alabed et al.}
\tolerance=2000 \hbadness=2000
\begin{abstract}
    \normalsize
    RocksDB is a general-purpose embedded key-value store used in multiple different settings.
    Its versatility comes at the cost of complex tuning configurations.
    This paper investigates maximizing the throughput of RocksDB IO operations by auto-tuning ten parameters of varying ranges.
    Off-the-shelf optimizers struggle with high-dimensional problem spaces and require a large number of training samples.

    We propose two techniques to tackle this problem: multi-task modeling and dimensionality reduction through clustering.
    By incorporating adjacent optimization in the model, the model converged faster and found complicated settings that other tuners could not find.
    This approach had an additional computational complexity overhead, which we mitigated by manually assigning parameters to each sub-goal through our knowledge of RocksDB.
    The model is then incorporated in a standard Bayesian Optimization loop to find parameters that maximize RocksDB's IO throughput.

    Our method achieved x$1.3$ improvement when benchmarked against a simulation of Facebook's social graph traffic,
    and converged in ten optimization steps compared to other state-of-the-art methods that required fifty steps.
\end{abstract}
\small
\maketitle
\normalsize
\section{Introduction}
A high-dimensional optimization space is a common phenomenon in general-purpose systems as they have many parameters and objectives.
Here we investigate the optimization of RocksDB, a popular key-value store and the building block of many mission-critical systems.
For example, Facebook uses RocksDB as a backend to diverse systems with high-query, high-throughput, or low-latency requirements \cite{caoCharacterizingModelingBenchmarking2020}.
These widely different use-cases have proved difficult to design a one-size fit all system due to these challenges:
\begin{itemize}
   \item RocksDB \cite{rocksdb_params} has over thirty parameters.
         The sheer number of configurations is overwhelming to the user.
   \item The hardware directly impacts the user's choice of configurations.
         For example, configurations that are valid on SSD perform suboptimally on HDD \cite{rocksdbdocumentationsTuningRocksDBSpinning}.
   \item Each application has unique access patterns \cite{caoCharacterizingModelingBenchmarking2020}.
\end{itemize}

\begin{figure}[b]
   \centering
   \includegraphics[width=1\linewidth]{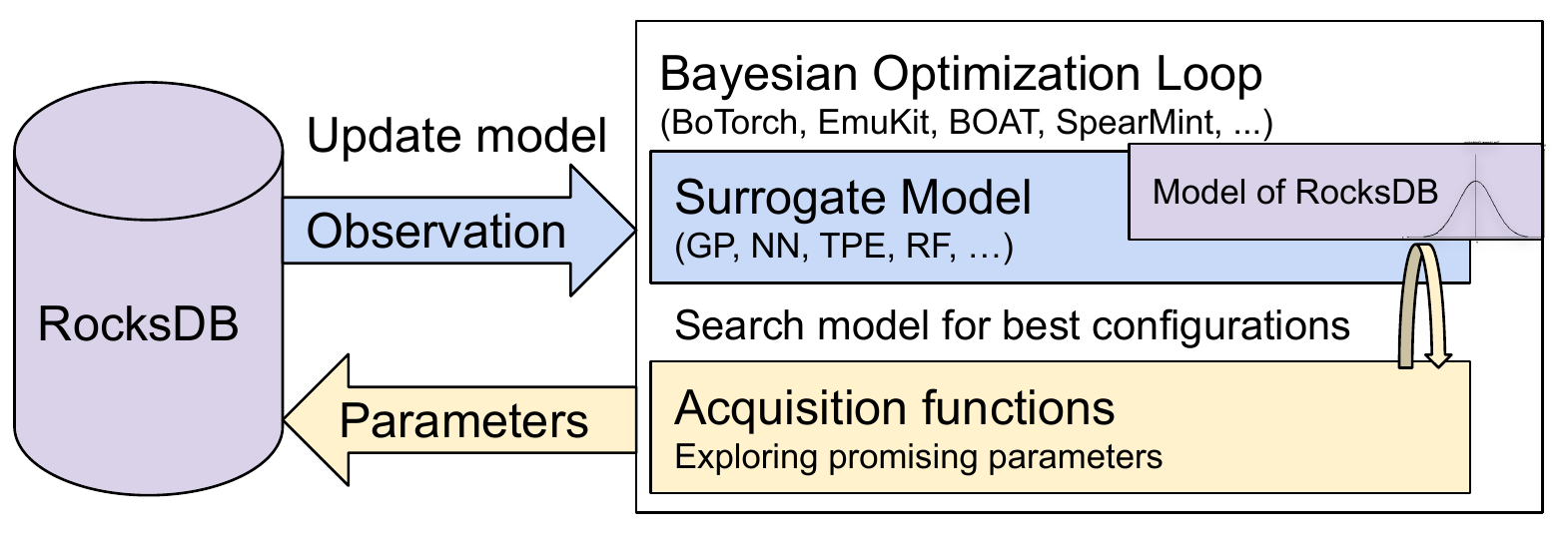}
   \caption[Bayesian optimization training loop.]{
      Bayesian optimization training loop. A surrogate model builds knowledge of the system.
      The acquisition function finds the most promising configurations in the model to try on the real system.
   }
   \label{fig:bo_loop}
   \Description{The figure illustrates a Bayesian optimization training loop.
      RocksDB reports its system metrics to the Bayesian optimization loop.
      The surrogate model updates its belief about the system.
      The acquisition function then performs a series of numerical optimization steps to find the most promising configurations in the model to try on the real system.
      The promising configuration is then used in the next spin-up of RocksDB.
   }
\end{figure}

These challenges motivate the need for auto-tuning methods.
Methods such as random search \cite{bergstraRandomSearchHyperparameter2012}, hill-climbing \cite{Ansel2010}, reinforcement learning \cite{suttonReinforcementLearningIntroduction1998},
or population-based search \cite{jaderbergPopulationBasedTraining2017} find optimal configurations in high-dimensions.
However, they suffer from requiring many training samples that are expensive to come by.
Every training sample requires a full instance restart and execution of a resource-intensive benchmark.

\gls{bo} \cite{snoekPracticalBayesianOptimization2012}, a sample efficent tuner, has received considerable attention in recent years due to its versatility and efficiency.
The framework, illustrated in \autoref{fig:bo_loop}, first builds a system model and then uses the model to find an optimal configuration, thus reducing the interaction with RocksDB and leading to fewer evaluations.

\gls{bo}'s drawback is its inability to handle high-dimensional spaces due to the \textit{curse of dimensionality} \cite{verleysenCurseDimensionalityData2005} and a computationally expensive operation in the surrogate model.
In this paper, we mitigate these drawbacks by adding expert knowledge to the \gls{bo} surrogate model that exploits the system's properties.
This work addresses the high-dimensionality problem with two novel ideas: optimizing over multiple targets to increase learning mileage per training sample; and sub-task decomposition to reduce the problem's dimensions.

\begin{figure}
   \centering
   \begin{subfigure}[b]{.225\textwidth}
      \includegraphics[width=1\linewidth]{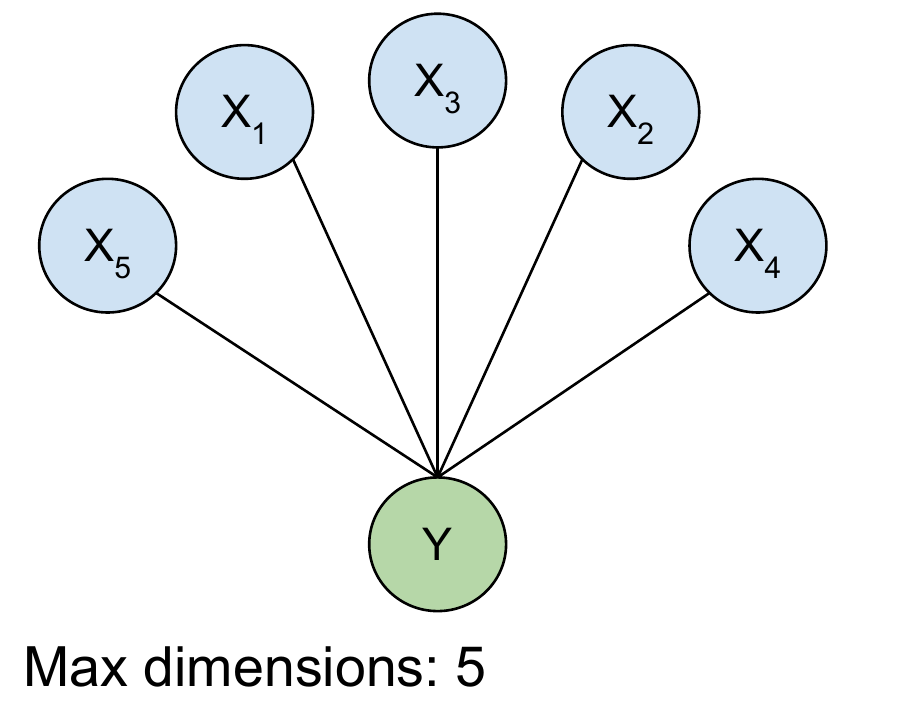}
      \caption{Without decomposition all parameters are considered.}
      \label{fig_decomposed_a}
   \end{subfigure}
   \hfill
   \begin{subfigure}[b]{.225\textwidth}
      \includegraphics[width=1\linewidth]{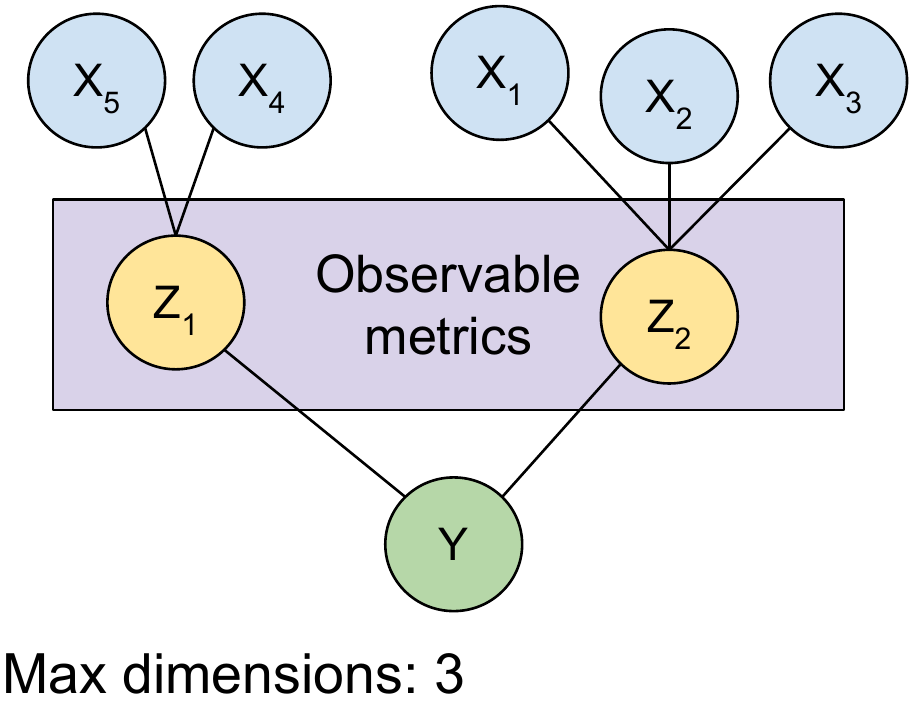}
      \caption{Decomposition reduces the considered dimensions.}
      \label{fig:decomposed_b}
   \end{subfigure}
   \vspace{3mm}
   \caption{Illustration of structural decomposability. Structural modeling reduces the dimensional space of the problem.
      \autoref{fig:decomposed_b} the maximum effective dimension is three for the the metric with largest number of parameters:$p(z_2|x_1, x_2, x_3)$.
   }
   \label{fig:decomposition}
   \Description{
      The Figure showing system decomposability.
      On the left, a standard optimization approach that considers all parameters jointly.
      On the right, a decomposed method breaking down optimization into a series of sub-tasks, leading to fewer dimensions considered for each optimization target.
   }
\end{figure}

Optimizing over multiple targets is especially beneficial to computer systems as their performance depends on various internal components at work.
For example, in the RocksDB case, the write throughput is bottlenecked by IO stalls, reducing the time spent stalling improves the throughput.
This knowledge provides the model with a broader overview of the system's behavior.

Adding optimization targets has an increased computational cost.
Hence the second part of this work mitigates this cost through decomposition.
\autoref{fig:decomposition} demonstrates structural decomposability: rather than jointly considering all parameters and assuming they all influence the main optimization target,
the primary optimization target decomposes into a subset of system components that influence the primary target.

We position our work as a contribution towards utilizing computer systems structure in an accessible way and
highlight it in a case-study of maximizing RocksDB's IO throughputs on Facebook's social graph workload \cite{caoCharacterizingModelingBenchmarking2020}.

In summary, our main contributions are the following:
\begin{itemize}
   \item A method to inject structural knowledge in a \gls{bo} optimizer significantly reducing the convergence time from seventy to ten steps.
   \item The reduction of the problem's dimensionality using manual task decomposition.
   \item An IO operations throughput improvement of up to x$1.45$ over RocksDB's default configurations.
\end{itemize}
\section{Background}
\subsection{RocksDB}
RocksDB is a key-value store based on LevelDB \cite{dentGettingStartedLevelDB2013} that provides efficient concurrent reads and writes.
RocksDB stores new data in a log-structured merge-tree format \cite{oneilLogstructuredMergetreeLSMtree1996} in memory (memtable).
Once the memtable is full, RocksDB flushes it sequentially into a Sorted Sequence Tree (SST) file on disk for persistent storage.
RocksDB still processes concurrent writes during the flushing process; hence, it is a popular choice for high-throughput applications such as stream processors \cite{carboneApacheFlinkStream2015}.

SST organizes the data in levels starting from level-0 to level-n. When a level is full, it performs a housekeeping operation (compaction) that merges levels and removes tombstones. During compaction, RocksDB stalls new writes, leading to a reduced IO throughput and increased latency, which is why each level's size is an important parameter.
A larger level size reduces compactions frequency; however, read operations will take longer to perform as there are more SST to scan before finding keys.
Readers are referred to \cite{dongOptimizingSpaceAmplification2017, facebookRocksDBHomepage2021} for a detailed information about RocksDB architecture.

\subsection{Bayesian optimization}
\gls{bo} is a sample efficient optimization framework \cite{snoekPracticalBayesianOptimization2012} that solves the problem of black-box function optimization.
It performs a series of iterative operations that builds knowledge of the function and then optimizes it.
Formally, it is expressed as $x^* = argmax(f(x))$,  where $f$ is the objective function that typically takes $x \in R^D$ as a parameter, and $D$ its dimensionality.

BO's algorithm shown in \ref{algo_bo} has two main components: a surrogate model and an acquisition function.
The surrogate model contains a belief of the system and updates at every new observation.
The acquisition function performs numerical optimization operations over the model to find configurations to test next on the objective function.
It aims to balance the trade-off between exploring and exploitation.
Exploring the system provides more information for the surrogate model while exploiting it to meet the primary optimization target.
\citet{shahriariTakingHumanOut2016} provides an in-depth survey of BO and its components.

\vspace{-2mm}
\begin{algorithm}
    \caption{Bayesian optimization}
    \label{algo_bo}
    \begin{flushleft}
        \textbf{Input:} acquisition function $\alpha$,
        surrogate model $M$\\
    \end{flushleft}
    \begin{algorithmic}
        \State Initialize a list $D$ holding the observed data
        \While{there is budget}

        \State select next point $x_{n+1}$ to evaluate
        \State $x_{n+1} = \alpha(M, D)$

        \State submit $x_{n+1}$ to be evaluated next and observe results
        \State $y_{n+1} = system(x_{n+1})$

        \State augment the observed data with a new observation
        \State $D \cup (x_{n+1}, y_{n+1})$
        \EndWhile

    \end{algorithmic}
\end{algorithm}

\subsection{Surrogate model} \label{sc:surrogate_models}
\subsubsection{\gls{gp}}
A \gls{gp} \cite{rasmussenGaussianProcessesMachine2003} is a popular and powerful non-parametric model that captures the relation between complex variables in a multivariate normal distribution.
The \gls{gp}'s ability to regularize and inject priors over the objective function space allows it to capture complex models efficiently.

Formally expressed as $GP(\mu, K)$ where $\mu$ is the mean function, often set to a constant value (zero or the mean of the data) and $K$ is a covariance kernel.
While \gls{gp} is powerful and sample efficient, they struggle in high-dimensional settings $x \in R^D, D \geq 10$ \cite{wangBayesianOptimizationBillion2016}.
That is due to two main reasons: firstly, the GP analytical inference is computationally expensive, as it performs a $O(N^3)$ matrix inversion operation to calculate the log-likelihood; secondly, when the dimensionality increases, every training sample provide less information due to the \textit{curse of dimensionality} \cite{verleysenCurseDimensionalityData2005}.
Furthermore, a fundamental assumption of a \gls{gp} is that the parameter space form a multivariate Gaussian distribution.

Previous work has reduced the complexity of the \gls{gp}.
For example, GPyTorch \cite{gardnerGPyTorchBlackboxMatrixMatrix2019} uses efficient operations in a matrix-to-matrix structure to improve the average complexity to $O(N^2)$ when there is a large number of training samples.
Other methods use approximate inference by backpropagating gradients and performing variational inference \cite{hensmanGaussianProcessesBig2013}, these methods are highly scalable and benefit from the Autograd optimizations provided by PyTorch and TensorFlow.
However, firstly, they require more training samples as they approximate the \gls{gp} to the nearest Gaussian; secondly, they lack eventual convergence guarantees.

\subsubsection{Alternative surrogate models}
BO's framework is modular and allows many other surrogate models. We list some of the popular ones here:
\begin{itemize}
    \item Additive kernels \cite{duvenaudAdditiveGaussianProcesses2011} decompose a function into a sum of independent low-dimensional kernels operating on a subset of the input space, allowing the model to handle high dimensional parameter space.
    \item Tree-structured Parzen Estimator (TPE) model \cite{bergstraAlgorithmsHyperparameterOptimization2011} is designed to handle discrete variables out-of-the-box.
          It collects and scores previously observed parameter space into two kernel estimator densities.
          One is the most performant observation, and the other branch is everything else.
          The densities act as the surrogate model.
          TPE scales linearly with the data and handles categorical parameters but is less sample efficient than GPs \cite{snoekPracticalBayesianOptimization2012}.
    \item Random forest (RF) \cite{liawClassificationRegressionRandomForest2002} is a model that linearly scales with the data and handles any parameter type.
          The acquisition function requires an uncertainty estimate from the surrogate model to perform exploration over its space.
          HyperMapper \cite{nardiAlgorithmicPerformanceAccuracyTradeoff2017} and SMAC \cite{hutterSequentialModelbasedOptimization2011} use RF as a surrogate model by providing the empirical variance as an uncertainty estimate.
\end{itemize}

\section{Structured multi-task optimization}
\begin{figure}[!b]
    \vspace{-2mm}
    \centering
    \includegraphics[width=1\linewidth]{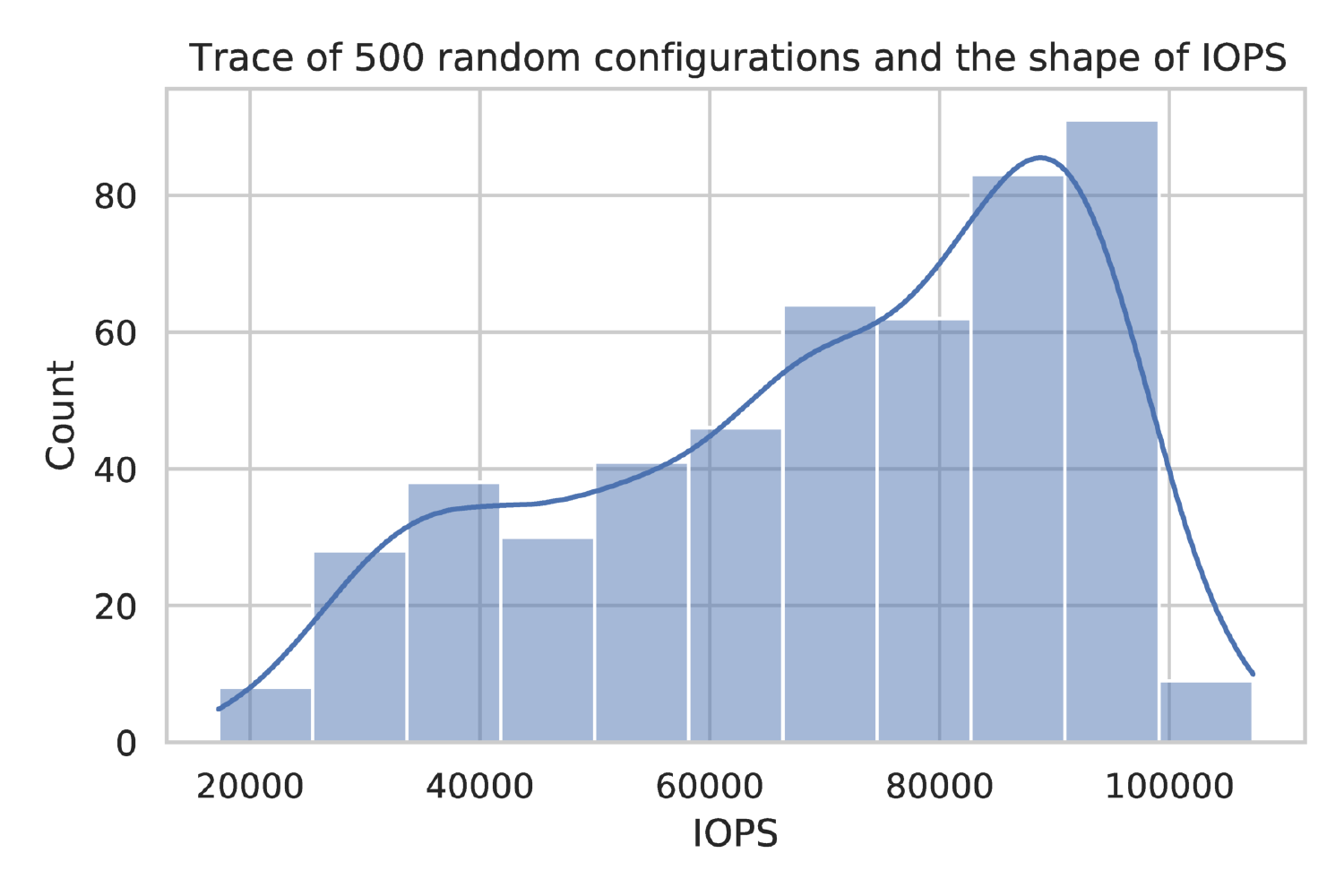}
    \caption[RocksDB IOPS shape randomly.]{
        Histogram of RocksDB's IO throughput when changing parameters randomly in $500$ samples is a smooth function that a Gaussian Process can approximate.}
    \label{fig:random_iops}
    \Description{Figure showing $500$ samples optimized using random optimization, the figure shows the observation follows a smooth function that a Gaussian Process can approximate.}
\end{figure}

\subsection{Overview}
Reducing the number of steps the tuner performs for finding optimal settings in RocksDB leads to considerable costs saving; hence it is the primary focus of this work.

As the previous section discussed, Bayesian optimization (BO) is a samples-efficient framework that has found successes in tuning systems.
Injecting experts' knowledge in BO is the subject of several interesting previous works \cite{neiswangerProBOVersatileBayesian2019,dalibardFrameworkBuildBespoke2016, nardiAlgorithmicPerformanceAccuracyTradeoff2017}.
While very performant and offering great flexibility, these methods have a steep learning curve.

This work utilizes an alternative mechanism to inject expert knowledge into the model;  the user identifies low-level metrics to optimize that also improve their main objective.
We used multi-task learning to capture the interaction between system components and learn more from every sample, reducing the observations needed for convergence.
We presented an extension that reduces the dimensions through a manual grouping of parameters to speed up the convergence by reducing the maximum dimension.

\subsection{Problem space and assumptions}
The fundamental assumption in using \gls{gp} is that the modeling space is a multivariate Gaussian distribution, and the function is differentiable at every point (smooth).
We performed $500$ independent experiments where we randomly sampled the modeling space (ten parameters) to learn if a GP can model the configurations' impact on the system.

\autoref{fig:random_iops}, shows a histogram of RocksDB's IO throughput (IOPS) when randomly sampling the selected ten parameters.
The histogram shows a smooth distribution that a \gls{gp} can model.
We do not consider categorical parameters as they cause jumps in the observed distribution, violating the smoothness assumption; this is a known issue for \gls{gp} models and outside our scope.

\subsection{Multi-task learning}
\subsubsection{Tasks.}
\begin{figure}
    \centering
    \includegraphics[width=1\linewidth]{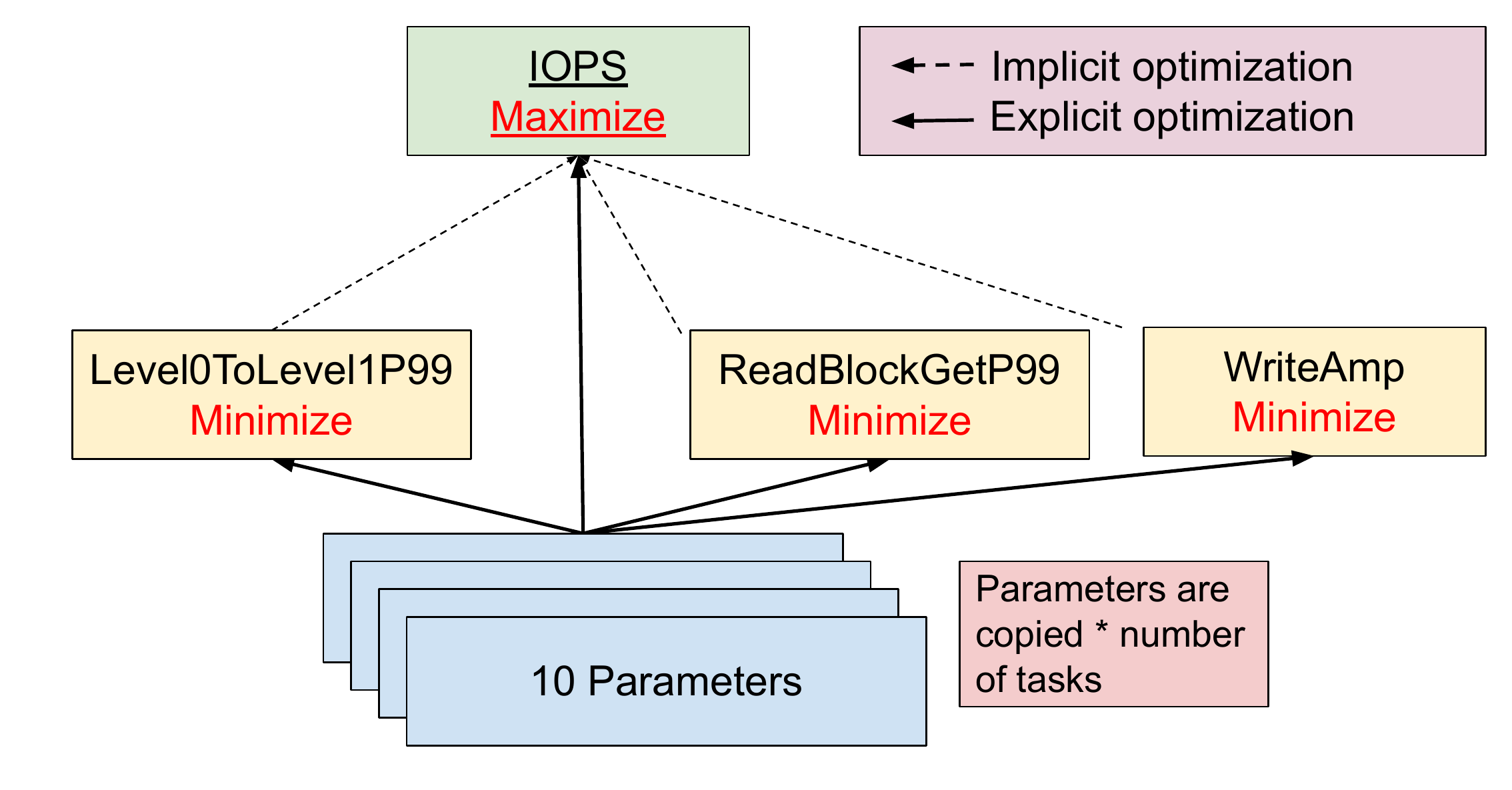}
    \caption[RocksDB multi-task optimization.]{
        A high-level view of multi-task optimization benefit in RocksDB. Optimizing the complimentary tasks provide implicit optimization to the primary goal, leading to TASK number of information per training sample.
    }
    \label{fig:multi_tasks}
    \Description{
        Figure showing a high-level view of multi-task optimization benefit in RocksDB. Optimizing the complimentary tasks provide implicit optimization to the primary goal, leading to TASK number of information per training sample.
    }
    \vspace{-4mm}
\end{figure}

This paper's thesis is that the main objective, IOPS, can be implicitly improved by optimizing adjacent objectives.
We chose three additional objectives based on our understanding of RocksDB architecture. Each of these objectives directly or indirectly impacts IOPS as illustrated in \autoref{fig:multi_tasks}.
These objectives each measures the performance of a separate component of RocksDB:
\begin{itemize}
    \item \textbf{WriteAmplification}: the ratio of bytes written to storage to the bytes written to the backend.
          Higher amplification reduces IOPS as it causes larger writes to disk and gets bottlenecked by the disk write speed.
    \item  \textbf{ReadBlockGetP99}: The $99th$ percentile latency to read a block of data.
          This target forces the optimizer to balance improving write throughput against read throughput, which provides a better IOPS.
    \item \textbf{Level0Tolevel1P99}: The $99th$ percentile time it takes to compact blocks stored in level0 to level1.
          IO stalls while the system is compacting the blocks. Thus this encodes the tradeoff between increased IOPS by using larger memory against having to stall later while compacting blocks out of the memory once it is full.
          Level0To1 is the most important one, as higher levels are stored on disk and larger, so they happen less regularly, while level0 flushes happen very frequently.
\end{itemize}

\subsubsection{Intrinsic coregionalization model.}
Modeling the three additional tasks requires changes to the kernel function of the \gls{gp}.
The popular method to enable multi-task learning in \gls{gp} uses the Intrinsic Coregionalization Model (ICM) \cite{alvarezKernelsVectorvaluedFunctions2012} kernel.
We used GPyTorch \cite{gardnerGPyTorchBlackboxMatrixMatrix2019} implementation of ICM kernel.
It is expressed in \autoref{math:icm}, where $k_x(x,x')$ is the parameter covariance kernel, and $k_T(m,m')$ is the task similarity kernel.
\begin{equation}\label{math:icm}
    k((x,m), (x',m')) = k_x(x,x')k_T(m,m')
\end{equation}
Using this kernel in the \gls{gp} outputs a multivariate Gaussian distribution for each of the tasks.
Each output is then optimized jointly as part of the \gls{bo} loop.

The ICM kernel is an excellent tool to learn implicit system structure as it learns two things: the covariance between configurations and observation, and a task similarity matrix for sharing knowledge between tasks.
The ability to share knowledge between tasks and provide $(TASKS*N)$ additional training samples for the model to utilize makes ICM suitable for improving its convergence speed.

Algorithm \ref{algo:multi_tasks} shows the preprocessing steps to using ICM for RocksDB tuning.
We scale every historical configuration used to tune RocksDB by the number of tasks available and normalize the configurations into a unit cube $[0, 1]^D$, so the model considers all parameters equally.
Next, tasks are standardized around zero mean and unit variance; this smoothens the task distribution's shape providing an easier target for the \gls{gp} to model.

\begin{center}
    \begin{algorithm}[H]
        \caption{RocksDB Multi-Task Model}
        \label{algo:multi_tasks}
        \begin{flushleft}
            \textbf{Input:} Observed metrics list $T$ and previous parameters $X$ \\
        \end{flushleft}
        \begin{algorithmic}
            \State Initialize $D$ linking parameters to their associated task.
            \State Normalize all previous observations $X$ to the unit cube.
            \State $\forall x \in X \Rightarrow x' = \frac{x - min(x)}{max(x)-min(x)}$
            \For{each \textit{task} $\mathit{t}$ in $T$}
            \State Standardize the task to a zero mean and a unit variance.
            \State $y_{t} = \frac{(\textit{task}) - mean(\textit{task})} {\sigma}$
            \State Assign previous observation to each task .
            \State $D \cup (x, y_{t})$
            \EndFor

            \State Model all the tasks using ICM kernel $K$
            \State \Return  $GP(D)$
        \end{algorithmic}
    \end{algorithm}
\end{center}

\subsubsection{ICM challenges}
ICM method provides a neat trick to get more mileage out of the few samples we usually have when tuning computer systems.
However, it raises some scalability concerns; a standard GP inference is $O(n^3)$, duplicating the data to the number of tasks scales this to $O(Tn^3)$ which can be troublesome.
While it can be argued that computer systems rarely have enough samples to reach that bottleneck, if we look into optimizing more than ten parameters, this will hit a wall very quickly.
This issue motivates the next subsection on decomposition.

\subsection{Decomposability through clustering}
\subsubsection{Decomposability in RocksDB}

The large parameter space in RocksDB is a challenge for most optimizers due to the \textit{curse of dimensionality} \cite{verleysenCurseDimensionalityData2005} and the high computational cost.
Using the multi-task method mitigates some of the first issues at an increased computational cost.

To reduce the computational cost, we exploit a fundamental property in RocksDB design; the final observed metric is the sum of multiple internal components performance; this is known in the literature as functional decomposability \cite{boydNotesDecompositionMethods2007}.
The decomposability refers to the smallest unit of observable RocksDB's performance metric and a corresponding set of parameters in this context.
By decomposing it, we reduce the effective dimensions as each model is self-contained, and thus the joint probability distribution of that model is much smaller.

\subsubsection{Manual clustering}

\begin{figure}
    \centering
    \includegraphics[width=1\linewidth]{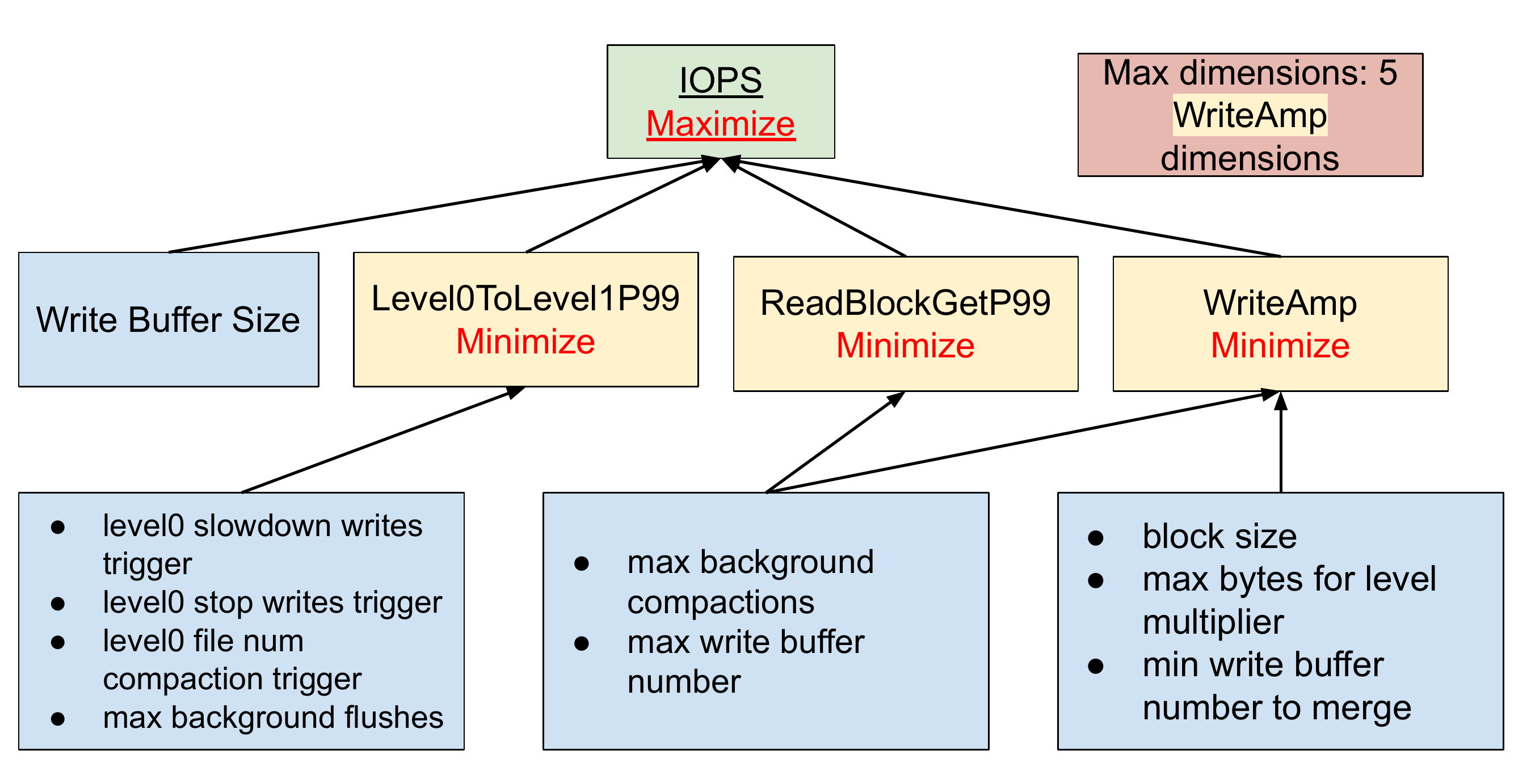}
    \caption[RocksDB clustered multi-task optimization.]{
        We manually assign a subset of parameters to each optimization target, thus reducing the dimensions.
        The assignment was done by picking association with the highest covariance values.
    }
    \label{fig:clustered_tasks}
    \Description{
        Figure illustrating the idea behind manually decomposing the parameter space and assigning it as a dedicated metric to optimize for. By doing so, we reduce the effective dimensional space to the size of the largest cluster.
    }
    \vspace{-4mm}
\end{figure}

As the modeling complexity linked directly to the dimension of the parameter space,
by slicing the dimensions, we get a significant boost to both inference time and reducing the impact of \textit{curse of dimensionality}.
Using the $500$ random configurations trace, we calculated the correlations between IOPS and the $517$ observable metric from RocksDB and the correlations between them to the parameters.
This reduced the pool of candidate clusters to $30$, of which we applied expert knowledge and performing isolated experiments to reach the structure shown in the figure.

We decompose the parameter space and use independent GPs for each cluster of parameters. Some of these GPs use the ICM kernel to share knowledge between the related subset of tasks and parameters.
\autoref{fig:clustered_tasks} illustrates the main idea behind this method.
This method provides the best of both worlds: reduced dimensionality from clustering and increased system knowledge from the shared kernel.

\subsubsection{Unsupervised clustering}
We recognize the opportunity to employ unsupervised clustering to find these clusters.
An unsupervised approach provides the most straightforward interface for the end-user;
the user provides the various metrics to the tuner, while the tuner takes on the responsibility to find structure in the system using these system observations.
The unsupervised approach combined with multi-task learning allows for learning in situations where a parameter is assigned to multiple tasks simultaneously.
A parameter impacting multiple different clusters can often happen in RocksDB.
For example, the buffer size has a positive correlation to latency but a negative correlation to compaction time.
We leave this as future work and show our preliminary results here.

\section{Evaluation}
\subsection{Setup}
We tuned ten RocksDB parameters to maximize IO operations throughput (IOPS).
Critically for an embedded key-value like RocksDB to be efficient when performing its IO operations \cite{dongOptimizingSpaceAmplification2017, ouaknineOptimizationRocksdbRedis2017}.
Maximizing IOPS yields improvement to the system regardless of the application characteristic.
\autoref{table_rocksdb} has descriptions and ranges of the parameters.
We initialized our \gls{bo} models with a single random observation to highlight our models' efficiency.

\begin{table}
    \begin{center}
        \small
        \caption{RocksDB parameters and their impact. All reported parameters are discrete ordinal variables.}
        \label{table_rocksdb}
        \vspace{2mm}
        \begin{tabular}{|l|l|l|}
            \toprule
            Parameter                              & Range          & Default  \\
            \midrule
            max\_background\_compactions           & $[1, 2^8]$     & $1$      \\
            max\_background\_flushes               & $[1 10]$       & $1$      \\
            write\_buffer\_size                    & $[1, 15*10^7]$ & $2^{26}$ \\
            max\_write\_buffer\_number             & $[1, 2^7]$     & $2$      \\
            min\_write\_buffer\_number\_to\_merge  & $[1, 2^5]$     & $1$      \\
            max\_bytes\_for\_level\_multiplier     & $[5, 15]$      & $10$     \\
            block\_size                            & $[1, 5*10^5]$  & $2^{12}$ \\
            level0\_file\_num\_compaction\_trigger & $[1, 2^8]$     & $2^2$    \\
            level0\_slowdown\_writes\_trigger      & $[1, 2^{10}]$  & $0$      \\
            level0\_stop\_writes\_trigger          & $[1, 2^{10}]$  & $36$     \\
            \bottomrule
        \end{tabular}
    \end{center}
\end{table}

\subsubsection{Experiment goals}
\begin{itemize}
    \item Success criteria for the tuner is to \textbf{converge faster} and find the \textbf{most performant} IO throughput.
    \item Highlight the efficiency of cluster-based multi-task approach in exploiting system decomposability.
\end{itemize}

\subsubsection{Benchmark}
We used RocksDB's benchmark tool \textit{db\_bench} to model Facebook's social graph traffic per the parameters reported in \cite{caoCharacterizingModelingBenchmarking2020}.
It runs $50$ million queries in fifteen-minutes.
This workload has a mixture of all RocksDB operations: $78\%$ GET, $13\%$ PUT, $6\%$ DELETE, and $3\%$ Iterate, the workload patterns change every $5000$ operation reflecting real workload.
A detailed discussion of this benchmark is in \citet{caoCharacterizingModelingBenchmarking2020}, who also argues that this is more realistic benchmark than the popular YCSB \cite{cooperBenchmarkingCloudServing2010}.
We documented the benchmark command in the appendix \ref{appendix:benchmarks}.
Every experiment repeated five times, with the variance visualized in the plots.
\vspace{-4mm}
\subsubsection{Libraries}
We used GPyTorch (v1.3) \cite{gardnerGPyTorchBlackboxMatrixMatrix2019} for the \gls{gp} implementation.
BoTorch (v0.3) \cite{balandatBoTorchProgrammableBayesian2019} for the implementation of the \textit{Expected improvement} \cite{vazquezConvergencePropertiesExpected2010} acquisition function in our implementation.
We used Microsoft NNI v2.0 \cite{opensourceMicrosoftNni2020} implementations of TPE, Random, and GP(NNI) in the baseline.

\vspace{-2mm}
\subsubsection{Hardware}
Google Cloud general compute instance e2-standard-4 (4 CPUs, 16GB RAM).
\vspace{-2mm}
\subsubsection{Baselines}
\autoref{tab:baselines} documents the baselines we used to maximize RocksDB's IOPS.
\vspace{-4mm}

\begin{center}
    \small
    \begin{table}[!h]
        \caption{
            \label{tab:baselines} Alternative surrogate models as baselines.
            The background has a  short introduction to these methods \ref{sc:surrogate_models}.
        }
        \vspace{2mm}
        \begin{tabular}{ | p{79pt} | p{142pt} | }
            \hline
            \textbf{Method} & \textbf{Use case}                 \\
            \hline

            TPE \cite{bergstraAlgorithmsHyperparameterOptimization2011}
                            & Handles discrete parameters.
            \\
            \hline

            GP (NNI) \cite{opensourceMicrosoftNni2020}
                            & Standard $O(n^3)$ implementation.
            \\
            \hline

            Random \cite{bergstraRandomSearchHyperparameter2012}
                            & Low effective search dimensions.
            \\
            \hline

            Additive kernel \cite{duvenaudAdditiveGaussianProcesses2011}
                            & Low-dimensions decomposability.
            \\
            \hline

            Default
                            & RocksDB v6.17 default settings.
            \\
            \hline

            BoTorch \cite{balandatBoTorchProgrammableBayesian2019}
                            & Efficient GPyTorch $O(n^2)$ GP.
            \\
            \hline
        \end{tabular}
        \vspace{-4mm}
    \end{table}
\end{center}

\subsection{IOPS performance}
\begin{figure}[H]
    \centering
    \includegraphics[width=1\linewidth]{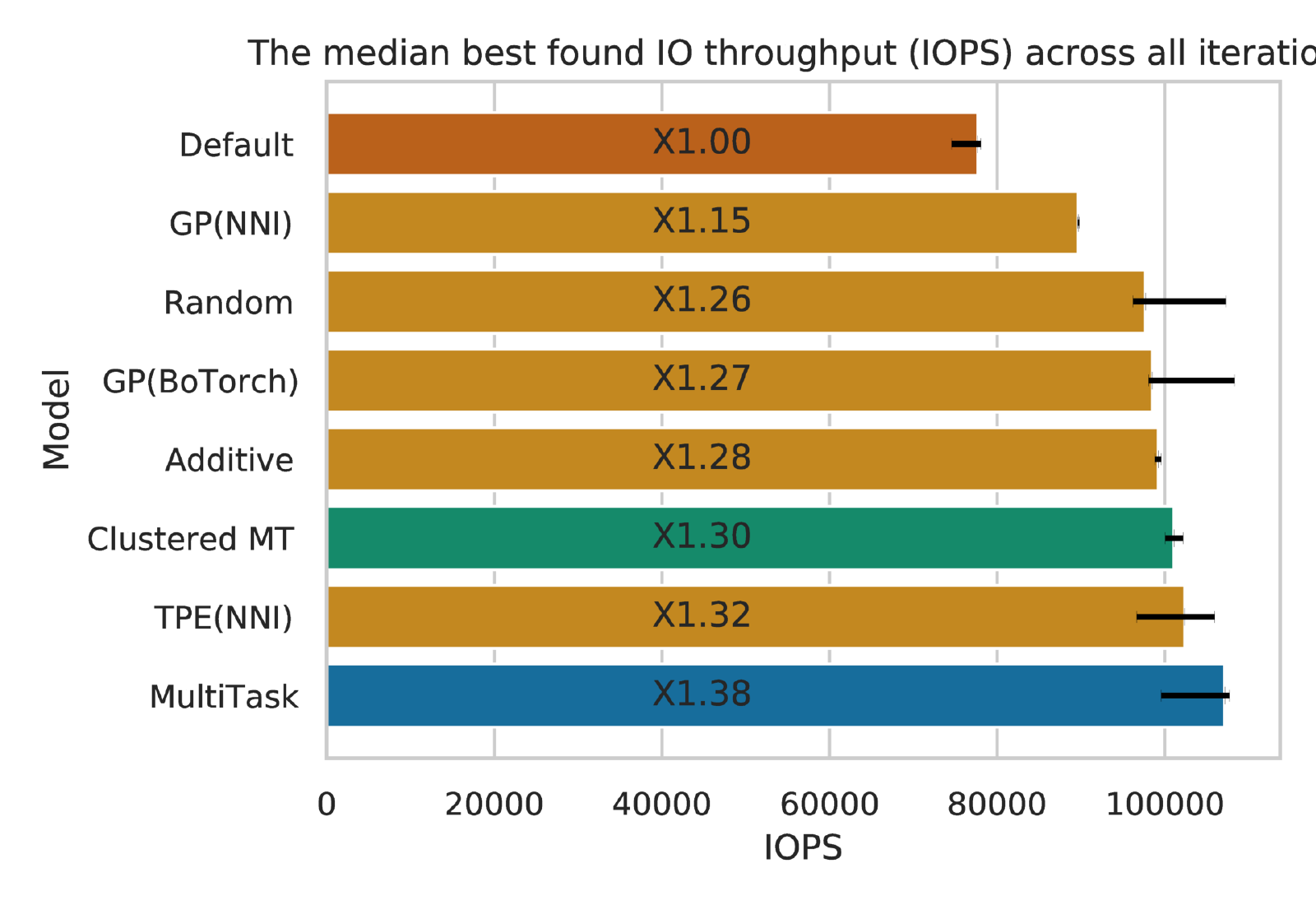}
    \vspace*{-6mm}
    \caption[Model to the best configuration found maximizing IOPS.]{
        The best IO throughput found in $100$ steps with the median of five runs reported with error bars showing the minimum and maximum achieved IOPS.
        The higher the IO throughput, the better.
        The middle text calculates the ratio of improvement over the default.
    }
    \Description{Figure is showing the best found IO throughput for every model.
        The best performant models are both multi-task. They outperform default by x1.45, achieving a throughput of $110,000$
    }
    \label{fig:iops_bar}
\end{figure}

We are interested in finding the best set of parameters that maximizes IOPS given a limited computational budget.
We set a budget of $100$ optimization steps to perform parameter search and reported the best-found IOPS in \autoref{fig:iops_bar}.

The multi-task approach utilized the additional knowledge and found a configuration that other tuners could not find given the budget.
The multi-task approach achieved a $x1.38 \pm 0.06$ increased throughput than RocksDB default settings.
While Clustered multi-task achieved comparable results to other tuners, it consistently achieved them and achieved them fast. This is visible in the next subsection.
The other baselines showed comparative optimization results, which leads us to investigate the next important metric, convergence speed.
\subsection{Convergence}
\begin{figure}[ht]
    \centering
    \includegraphics[width=1\linewidth]{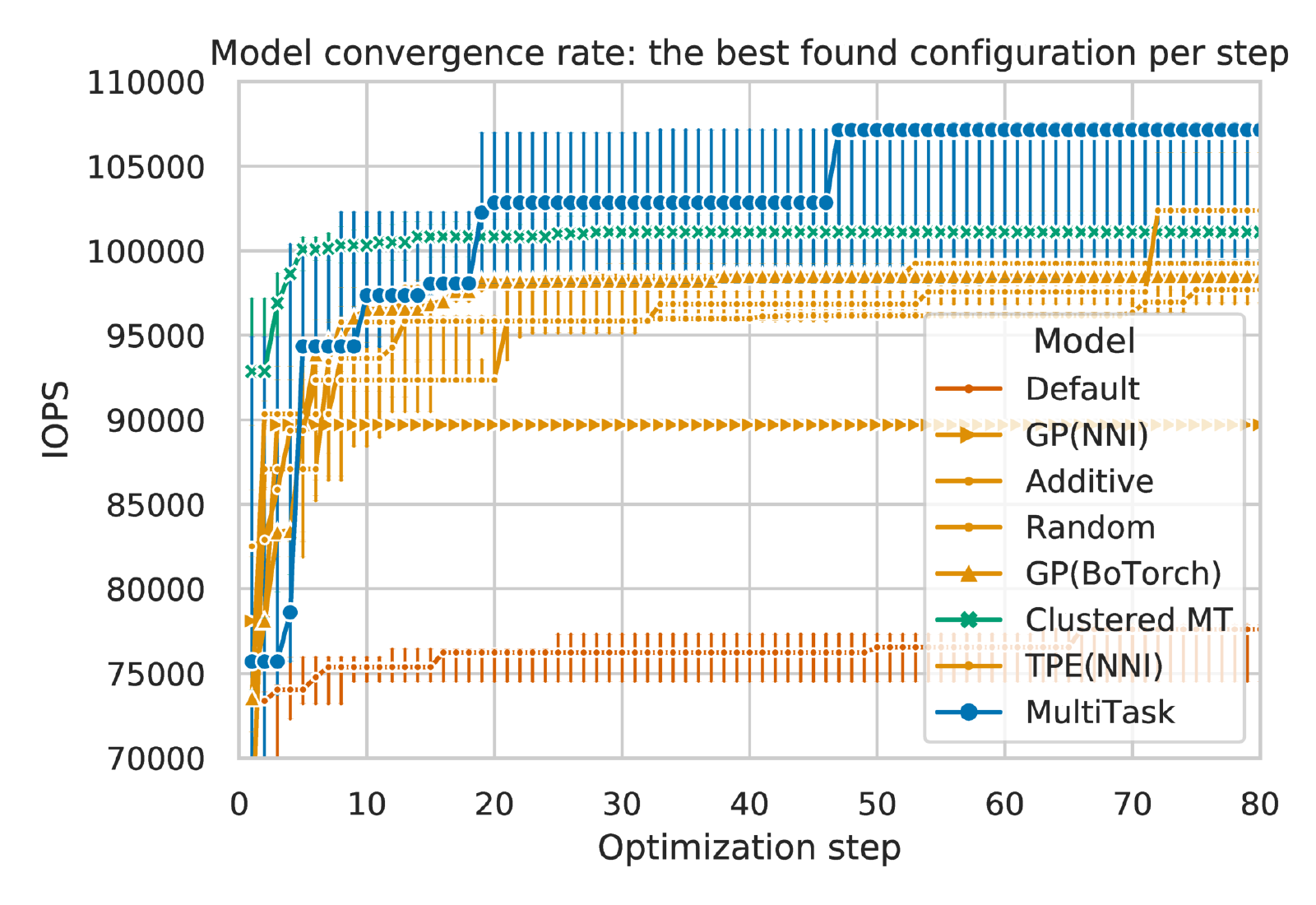}
    \caption[Model convergence speed towards best IOPS.]{
        This plot reports the best configurations the tuner finds in every step.
        The X-axis shows the number of steps the optimizer took.
        The Y-axis represents the median best-found configuration.
        The error bar shows the best and worst run in five benchmark runs.
        The higher the IOPS, the better.
        The figure is cut to $80$ steps since nothing changes after.
        We show a larger figure in the appendix \autoref{fig:convergence_colorful}.
    }
    \label{fig:convergence}
    \Description{Figure showing the convergence speed of every model, it can be observed that the structured multi-task able to find good performing parameters within the first ten steps, while the multi-task is the second-fastest converging method finding optimal parameters in 40 step, it took every other tuner 80 steps to find sub-optimal parameters.}
    \vspace{-2mm}
\end{figure}

An essential property of a tuner is to find the best configurations quickly.
Computer systems evaluation is expensive to come by, and the computational budget of $100$ we set is too generous.
Here we evaluate each model convergence speed.

\autoref{fig:convergence} shows the rate of convergence.
Both multi-task and clustered multi-task (\textit{Clustered MT}) approaches can find better configurations and find them sooner than the baselines.
It took \textit{Clustered MT} $10$ iterations consistently to find configurations that provide $x1.3$ improvement over the default, and it took the next non-cluster-based surrogate model $60$ iterations to reach comparable results.
However, the non-clustered multi-task approach was able to find better configurations (x$1.38$) in the long run (sixty steps) but was unstable where some experiments had worse performance.

It is worth noting \textit{GP(NNI)'s} flatline.
The standard \gls{gp} implementation started throwing memory exceptions and timeout due to its inability to handle the ten parameters.
We show a more efficient implementation of \gls{gp} in \textit{GP(BoTorch)} which utilizes GPyTorch \cite{gardnerGPyTorchBlackboxMatrixMatrix2019} as discussed in the background \ref{sc:surrogate_models}.

In summary, we showed that by providing the tuner with additional context of the system, it could model the search space more efficiently than other models.

\section{Related work}
\textbf{Structured optimization.}
Both BOAT \cite{Dalibard2017} and ProBO \cite{neiswangerProBOVersatileBayesian2019} demonstrated that adding structure to the surrogate model of the \gls{bo} loop provides significant convergence speedup.
BOAT allows the user to define semi-structured models: parametric models to capture trends and \gls{gp} to generalize and aggregate.

This work distinguishes itself by providing a straightforward mechanism for injecting expert knowledge by exploiting correlated targets through MultiTask learning.
We could not evaluate against them as ProBO is not publically available, while BOAT requires substantial work to integrate. 

\textbf{Multi-fidelity optimization.}
Multi-task learning has many similarities to the multi-fidelity literature \cite{kandasamyMultifidelityBayesianOptimisation2017}.
Both approaches are identical for all intended purposes,
as in a multi-fidelity approach with an exact number of observations for every objective, it will use an ICM kernel similar to the standard multi-task approach.
The only key difference is the philosophy of optimization.

In the multi-fidelity framework, the low-fidelity targets are cheap to evaluate objectives that the user does not care about (in this paper, it would be the three adjacent objectives). The primary objective is modeled as the high-fidelity objective and a function of the lower fidelities.
Since in this paper we actively chose the secondary objective, from the user's perspective, these secondary objectives are essential to optimize as well, which makes the problem better posed as multi-task learning rather than multi-fidelity.

\textbf{Multi-objective optimization}(MOO) \cite{blankPymooMultiobjectiveOptimization2020} shares the philosophy of our work.
In MOO, a Pareto frontier is calculated for all objectives, and the optimizer aims to balance the trade-off between optimizing the objectives according to the user's preference.
Our work requires the user to identify dependent tasks, so the complex trade-off must be calculated as the user wants to maximize them all equally.

\textbf{Reinforcement learning.}(RL) \cite{suttonReinforcementLearningIntroduction1998} successfully applied for computer system optimization \cite{Schaarschmidt2018, Pavlo2017b, Sharma2018, rlDevicePlacement}.
It has the advantage of handling a large number of parameters efficiently and is extremely easy to use. However, it suffers from several drawbacks:
first, it is susceptible to initializing seed due to the large number of stochastic processes used internally; this problem discussed in details in \cite{hendersonDeepReinforcementLearning2017};
secondly, it requires many training samples; thirdly, its results are uninterpretable and hard to debug; and finally, it is challenging to include expert knowledge in the model.

Our approach protects against these issues: \gls{gp} are resilient against the impact of a bad seed \cite{rasmussenGaussianProcessesMachine2003}.
The multi-task structure provides an interpretation of the model performance to the system expert and easy to understand mechanism to inject their knowledge into the model.
Finally, the multi-task structure provides each training step with many more samples, leading to faster convergence.

\textbf{Self-tuning databases}
This work can fall under self-tuning databases literature.
\citet{schmiedGeneralFrameworkMLbased2020} proposes a pipeline of machine learning techniques to tackle the large parameter space. The pipeline reduces the parameter space and selects the top ten promising parameters and their ranges to perform the \gls{bo} or RL loop to optimize the main objective.
Their work is orthogonal to our approach. The pipeline can be used as a preprocessing step to select more parameters to optimize and then utilize the multi-task approach to provide faster convergence.

\textbf{RocksDB optimizations.}
\citet{ouaknineOptimizationRocksdbRedis2017} tuned six RocksDB parameters manually to achieve x$11$ throughput.
While \citet{dongOptimizingSpaceAmplification2017} improved space and read efficiency by tuning different six parameters.
Both cases show the need for an auto-optimizer for RocksDB to achieve its potential.

\section{Conclusion}
This work presents an efficient auto-tuner for RocksDB that handles its high-dimensional space.
The tuner exploits alternative observable metrics and structural decomposability to converge faster and reduce the dimensional space.
We utilize multi-task learning to provide an accessible mechanism for expressing structure in the model.
The methodology is universal and can be applied to systems other than RocksDB as long as they expose their internal components metrics.

We demonstrated the tuner's effectiveness by tuning ten parameters to maximize IO throughput. 
The tuner outperformed the default configuration by x$1.35$ in $10$ iterations, compared to the other state-of-the-art methods requiring $60$ iterations.

\begin{acks}
  We would like to thank the reviewers for their valuable feedback.
  A special thanks to Thomas Vanderstichele for his comments that improved the readability of the paper.
  This research was supported by the Alan Turing Institute.
\end{acks}

\bibliographystyle{ACM-Reference-Format}
\bibliography{references}

\appendix
\section{Appendix A}
\tolerance=2000 \hbadness=2000

\subsection{RocksDB parameters}
We selected ten RocksDB parameters and here we summarize (or directly quote) RocksDB documentation \cite{rocksdb_params} on these parameters.

\subsubsection{MemTable Parameters}
\begin{itemize}
    \item \textbf{write\_buffer\_size}: RocksDB enables fast writes by inserting first into memtables (in-memory immutable data structure).
          Once the memtable is full, a new memtable is created.
          This parameter sets the size of that memtable. There is a complex tradeoff here between write rate and write duration and overall system performance.
    \item \textbf{min\_write\_buffer\_number\_to\_merge}: The created memtable live in memory until they reach this parameter value, then all memtables merged and flushed into a desk.
          Increasing this value reduces the space needed on the disk but reduces the read performance as every read needs to scan the memtable linearly.
    \item \textbf{max\_write\_buffer\_number}: Threshold that decides the number of write buffers held in memory impacts memory usage. The larger buffer will result in a longer recovery time.
\end{itemize}

\subsubsection{Block Cache}
\begin{itemize}
    \item \textbf{block\_size}: RocksDB stores data into a list of blocks. When a key is requested, a whole block associated with the key is loaded.
          A larger block size decreases memory usage and reduces wasted space (since there are fewer items to index and the index itself is smaller),
          but increases the number of reads needed to reach the data.
\end{itemize}

\subsubsection{Compaction}
\begin{itemize}
    \item \textbf{max\_background\_compactions}: is the maximum number of concurrent background compactions.
          A larger value maximizes CPU and storage.

    \item \textbf{max\_background\_flushes}: is the maximum number of concurrent flush operations.

    \item \textbf{max\_bytes\_for\_level\_multiplier}:
          A multiplier to compute max bytes for levelN $N \geq 2$,
          each subsequenet level is max\_bytes\_for\_level\_multiplier larger than previous one.
\end{itemize}

\subsubsection{levels parameters}
\begin{itemize}
    \item \textbf{level0\_file\_num\_compaction\_trigger}: RocksDB organizes its data in levels, where the newest data live in level0 and older data on lower levels.
          Level0 is unique because it may contain duplicates or deleted values, and thus the reads need to scan every file in level 0; this is not the case in successive levels.
          Moving files from level 0 to 1 requires a compaction procedure, and this is done on a single thread, and often it is the bottleneck in the system.
          A slow level0 of compaction halts the whole system since no other level can be scanned in the meantime.
          Therefore, this parameter is very complicated, and it is tough to derive useful intermediate statistics for its performance.
    \item \textbf{level0\_slowdown\_writes\_trigger}: Trigger to stall writes to DB writes once triggered.
    \item \textbf{level0\_stop\_writes\_trigger}: Trigger to stop writing completely once triggered.
\end{itemize}

\begin{figure*}
    \centering
    \includegraphics[width=\linewidth]{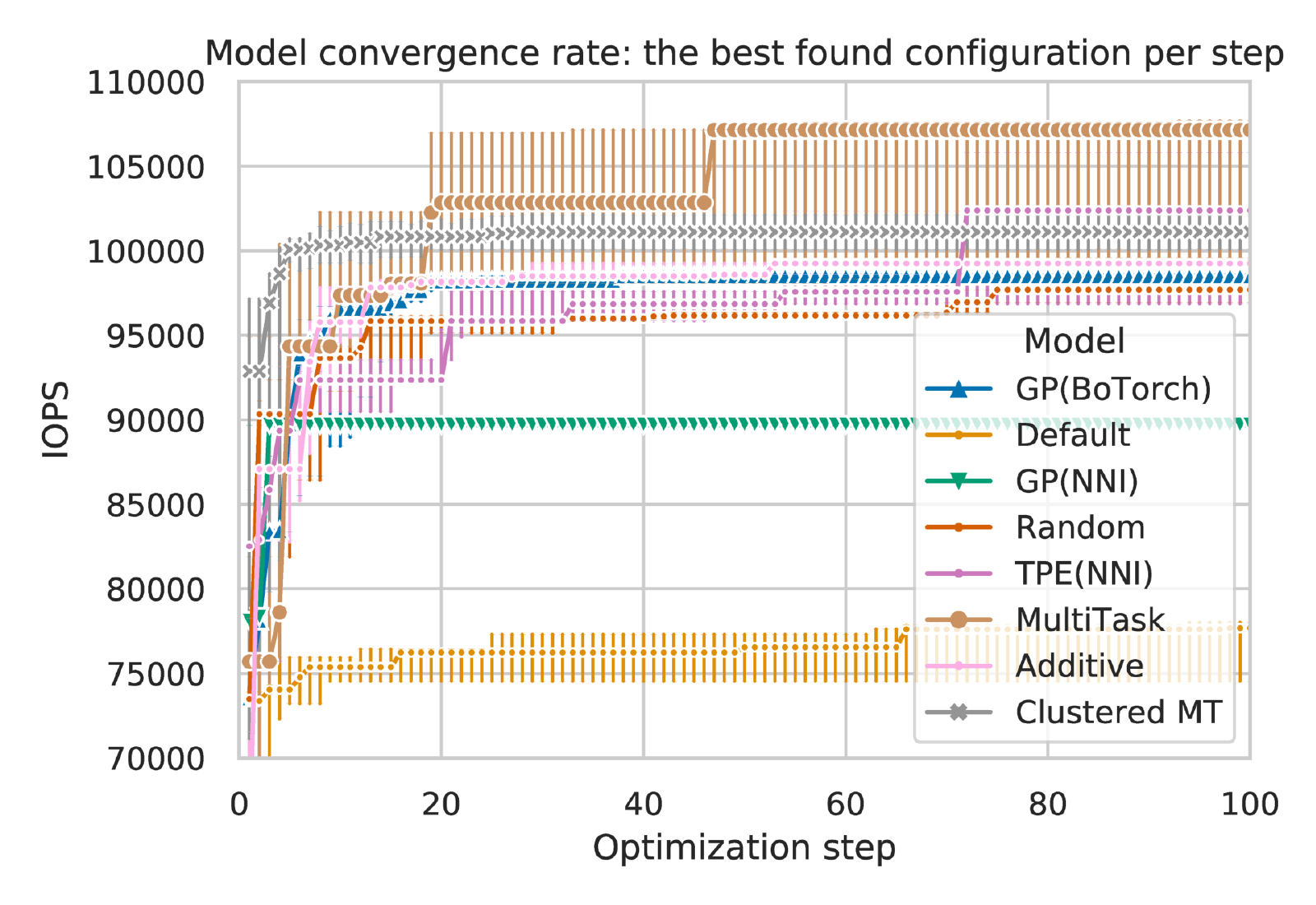}
    \caption[Model convergence speed towards best IOPS.]{
        A larger and more colorful plot than the one displayed in \autoref{fig:convergence} for easier readability.
        This plot reports the best configurations the tuner finds in every step.
        The X-axis shows the number of steps the optimizer took.
        The Y-axis represents the median best-found configuration.
        The error bar shows the best and worst run in five benchmark runs.
        The higher the IOPS, the better.
    }
    \label{fig:convergence_colorful}
    \Description{Figure showing the convergence speed of every model, it can be observed that the structured multi-task able to find good performing parameters within the first ten steps, while the multi-task is the second-fastest converging method finding optimal parameters in 40 step, it took every other tuner 80 steps to find sub-optimal parameters.}
\end{figure*}

\subsubsection{RocksDB stats file}
RocksDB stats file contains the following:
\begin{itemize}
    \item level0 to N compaction time and rate.
    \item stats of major and minor compaction of the SSTables, which is used to identify block size and write buffer impact.
    \item size of writes to older levels.
    \item number of records during compaction and number of duplicates and deletes found during compaction.
    \item Micro and Macro statistics of cache, block sizes, stalls, and access to each level.
\end{itemize}

\subsection{Benchmark commands}
\autoref{appendix:benchmarks} shows the command we used to run the benchmark.
This command replicates Facebook's social graph model reported in \cite{caoCharacterizingModelingBenchmarking2020}.
We increased the workload's intensity by 1000 times (increased sine\_d and reduced sine\_b) to reduce running the benchmark execution from $24$ hours to $15$ minutes.
This workload mixes all RocksDB operations in different ratios: $78\%$ GET, $13\%$ PUT, $6\%$ DELETE, and $3\%$ Iterate.

\begin{lstlisting}[caption={db\_bench execution command},label={appendix:benchmarks},language=Bash]
$db_bench --benchmarks="mixgraph,stats" -use_direct_io_for_flush_and_compaction=true -use_direct_reads=true -cache_size=26843 5456 -keyrange_dist_a=14.18 -keyrange_dist_b=-2.917 -keyrange_dist_c=0.0164 -keyrange_dist_d=-0.08082 -keyrange_num=30 -value_k=0.2615 -value_sigma=25.45 -iter_k=2.517 -iter_sigma=14.236 -mix_get_ratio=0.85 -mix_put_ratio=0.14 -mix_seek_ratio=0.01 -sine_mix_rate_interval_milliseconds=5000 -sine_a=1000 -sine_b=0.000000073 -sine_d=4500000 --perf_level=1 -reads=4200000 -num=50000000 -key_size=48 --statistics=1 --duration=300
\end{lstlisting}
\end{document}